\documentclass[10pt,twocolumn,letterpaper]{article}

\usepackage[pagenumbers]{cvpr} % To force page numbers, e.g. for an arXiv version
\usepackage{graphicx}
\usepackage{amsmath}
\usepackage{amssymb}
\usepackage{booktabs}
\usepackage{xcolor}
\usepackage{algorithm}
\usepackage{algpseudocode}

\usepackage{tcolorbox}
\tcbuselibrary{breakable,skins,listings}

\definecolor{cvprblue}{rgb}{0.21,0.49,0.74}
\usepackage[pagebackref,breaklinks,colorlinks,allcolors=cvprblue]{hyperref}

\def\paperID{10152} % *** Enter the Paper ID here
\def\confName{CVPR}
\def\confYear{2026}

\title{Agentic Video Intelligence: A Flexible Framework for Advanced Video Exploration and Understanding}

\author{
    Hong Gao$^{1,2}$\thanks{Equal Contribution.} \quad
    Yiming Bao$^{2}$\footnotemark[1] \quad
    Xuezhen Tu$^{2}$ \quad
    Yutong Xu$^{2}$ \quad
    Yue Jin$^{2}$  \quad
    Yiyang Mu$^{2}$ \\
    Bin Zhong$^{2}$ \quad
    Linan Yue$^{1}$ \quad
    Min-Ling Zhang$^{1}$\thanks{Corresponding Author.} \\
    $^{1}$SouthEast University, $^{2}$ZTE Corporation
}

\begin{document}
\maketitle
\begin{abstract}
Video understanding requires not only visual recognition but also complex reasoning. While Vision-Language Models (VLMs) demonstrate impressive capabilities, they typically process videos largely in a single-pass manner with limited support for evidence revisit and iterative refinement. While recently emerging agent-based methods enable long-horizon reasoning, they either depend heavily on expensive proprietary models or require extensive agentic RL training. To overcome these limitations, we propose \textbf{A}gentic \textbf{V}ideo \textbf{I}ntelligence (\textbf{AVI}), a flexible and training-free framework that can mirror human video comprehension through system-level design and optimization. AVI introduces three key innovations: (1) a human-inspired three-phase reasoning process (Retrieve-Perceive-Review) that ensures both sufficient global exploration and focused local analysis, (2) a structured video knowledge base organized through entity graphs, along with multi-granularity integrated tools, constituting the agent's interaction environment, and (3) an open-source model ensemble combining reasoning LLMs with lightweight base CV models and VLM, eliminating dependence on proprietary APIs or RL training. Experiments on LVBench, VideoMME-Long, LongVideoBench, and Charades-STA demonstrate that AVI achieves competitive performance while offering superior interpretability.
\end{abstract}
\section{Introduction}
\label{sec:intro}

\begin{figure}
    \centering
    \includegraphics[width=\linewidth]{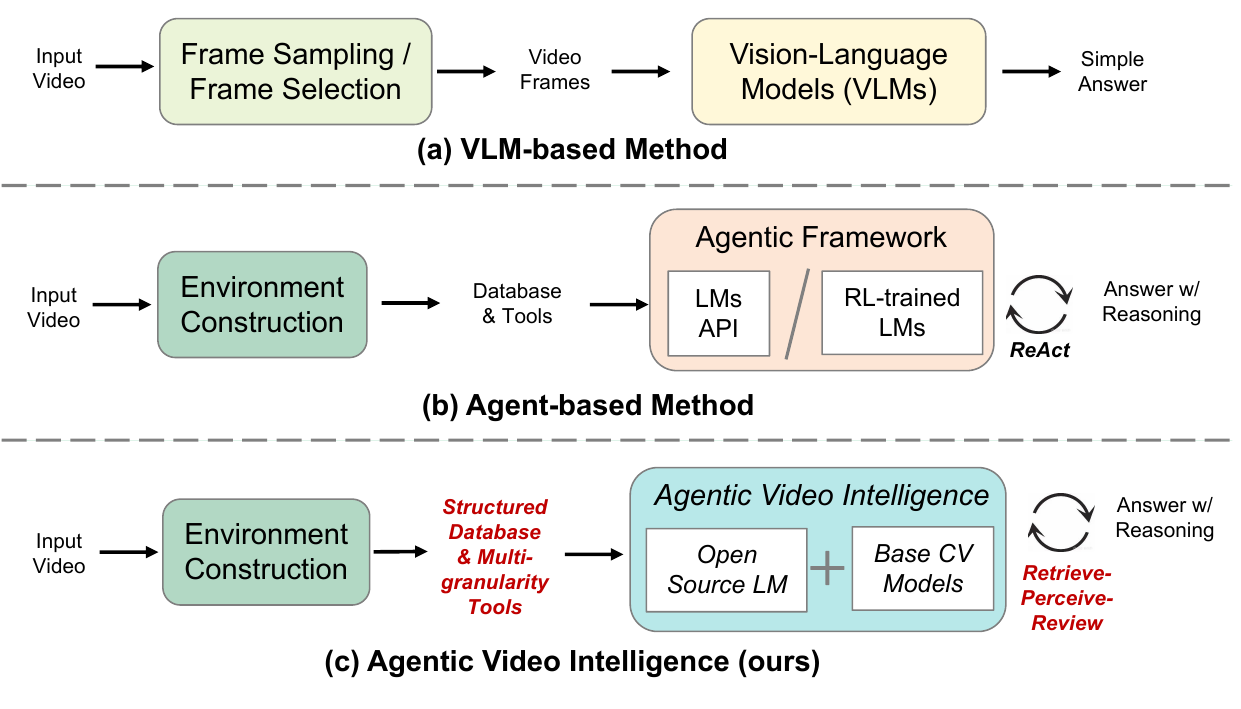}
    \caption{Comparison of video understanding paradigms. (a)VLM-based methods process video frames in a single-pass manner. (b) Current agent-based methods use ReAct loops to interact with environment, but requiring LMs API or RL training. (c) Our AVI imitates human intelligence with three-phase reasoning, builds structured environment, and employs open-source models.}
    \label{fig_intro}
\end{figure}

Understanding complex videos remains one of the most challenging tasks in computer vision. Unlike static images, videos encode temporal narratives, causal relationships, and multi-modal semantics that unfold across extended timescales~\cite{wang2025lvbench,fu2025videomme,wu2024longvideobench}. This complexity has motivated two dominant paradigms in recent research.

The first paradigm, as shown in ~\cref{fig_intro} (a), leverages Vision-Language Models (VLMs)~\cite{bai2025qwen2.5vl,zhang2025videollama3,wang2025internvideo2} to directly answer video understanding questions by processing video frames derived from uniform sampling~\cite{bai2025qwen2.5vl} or key-frame selection~\cite{tang2025adaptive-aks,gao2025apvr}. Although achieving impressive results, these methods lack interpretability and error correction mechanism. Instead, the monolithic VLMs are required to handle thousands of visual tokens in a single-pass manner. Without using external tools or seeking database, it struggles in simultaneously exploring temporal narratives and solving diverse video understanding tasks.

Recently, agent-based methods attempt to address these issues by decomposing video understanding into interpretable action sequences. As illustrated in ~\cref{fig_intro} (b), video agents iteratively interact with the pre-constructed environment (video database and executable tools) through reasoning action (ReAct~\cite{yao2022react}) loop, and then offer final answer as well as reasoning process and decision transparency. However, existing approaches face some critical barriers: (1) lack of structured video database that hinders the exploration of entity interactions~\cite{wang2024videoagent,zhang2024omagent},  (2) heavy reliance on expensive proprietary LM APIs (e.g., GPT-4o) that limits accessibility and reproducibility~\cite{zhang2025dvd,yuan2025videodeepresearch}, (3) alternative approaches demand resource-intensive agentic RL training that also reduces cross-task generalizability~\cite{long2025m3-agent,zhang2025vital,zhou2025reagent-v}.

To address these limitations, we propose \textbf{A}gentic \textbf{V}ideo \textbf{I}ntelligence (\textbf{AVI}), a flexible framework that mirrors human cognitive processes for advanced video exploration and understanding. As illustrated in ~\cref{fig_intro} (c), AVI comprises two synergistic stages: structured environment construction and three-phase agentic reasoning.

Specifically, drawing inspiration from human high-level structured memory, we extract abundant video information to build a textual, graph-based knowledge base. This construction occurs once per video, enabling efficient multi-question inference without repeatedly processing raw frames. The database organizes multi-granularity information including entity-centric graphs, captions of video clips and their embeddings. Moreover, we also design a toolkit of executable computer vision models for the agentic system to exploring database and the raw visual clues.

We further introduce a three-phase framework, \textit{\textbf{Retrieve-Perceive-Review}}, to iteratively explore global textual database or focus on local visual analysis through tool-integrated reasoning. Analogous to human dragging progress bars to gather contextual information, the \textit{Retrieve} phase identifies coarse-grained but sufficient candidate time ranges. Then mirroring humans zooming in to examine details, the \textit{Perceive} phase then deploys specialized CV tools for fine-grained visual analysis within some identified ranges. Finally, the \textit{Review} phase evaluates current understanding through reflection and decides whether to output the final report or return back to the \textit{Perceive} phase for refinement. Critically, this system builds upon open-source reasoning LLMs and lightweight base models, eliminating dependence on proprietary APIs or resource-intensive RL training through careful system-level design.

We conduct extensive experiments on three long video understanding benchmark including LVbench, VideoMME-Long, and LongVideoBench, as well as a temporal grounding benchmark Charades-STA. The results demonstrate that AVI achieves competitive performance while offering superior interpretability, cost-efficiency, and modularity. Our main contributions are as follows:

\indent $\bullet$ We propose Agentic Video Intelligence (AVI), a flexible and human-inspired framework with three-phase reasoning (Retrieve-Perceive-Review) that balances global exploration and local analysis for video understanding.

\indent $\bullet$ We introduce a structured and multi-granularity environment comprising an entity-centric knowledge base and executable CV tools, enabling efficient and interpretable tool-integrated reasoning with textual and visual seeking.

\indent $\bullet$ We demonstrate that open-source model ensembles—combining reasoning-capable LLMs with lightweight CV models—can achieve competitive performance without proprietary API dependence or RL training, significantly improving reproducibility and accessibility.

\begin{figure*}[!ht]
    \centering
    \includegraphics[width=\linewidth]{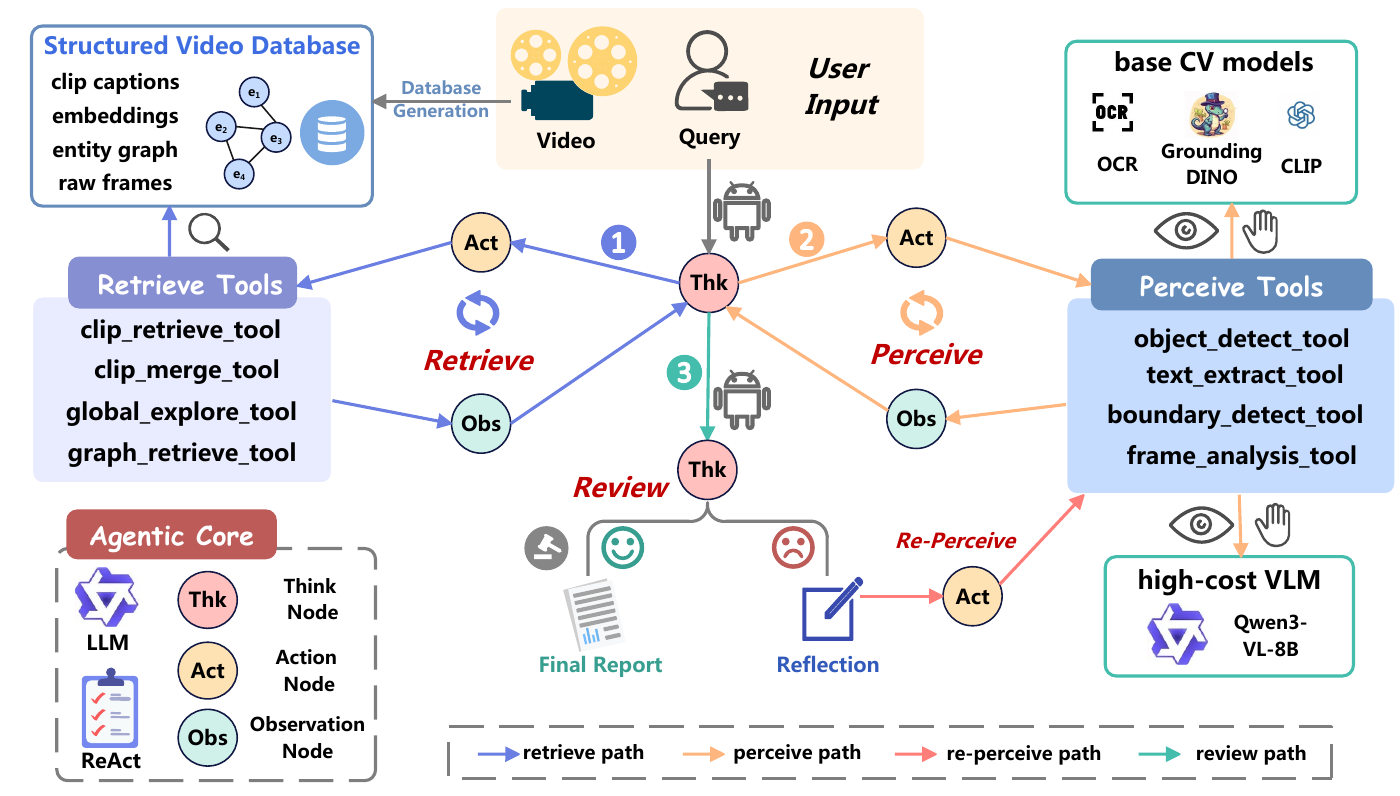}
    \caption{The AVI framework architecture. The structured video database contains clip captions, embeddings, entity graphs, and raw frames. The agentic core implements three-phase reasoning (Retrieve-Perceive-Review) with internal Think-Action-Observation nodes and two tool suites:(1) Retrieve Tools for global exploration and segment location; (2) Perceive Tools for local visual analysis, powered by base CV models and open-source VLM. Our AVI iteratively gathers evidence (Retrieve and Perceive phase) and determines whether to output or return back for refinement (Review phase). This design enables interpretable, training-free video understanding.}
    \label{fig_framework}
\end{figure*}
\section{Related Works}

\subsection{Video Understanding}
Recent advances in video understanding have been driven primarily by Vision-Language Models (VLMs)~\cite{bai2025qwen2.5vl,zhu2025internvl3,openai2023gpt4v}. These models extend large language models with visual encoders to process videos as sequences of visual tokens~\cite{lin2023videollava}. Representative works employ different temporal modeling strategies to balance efficiency and performance. Uniform sampling methods extract frames at fixed intervals, while adaptive selection approaches use attention scores or clustering to identify key frames~\cite{gao2025apvr,tang2025adaptive-aks}. Hierarchical methods process videos at multiple temporal granularities and reduce computational overhead by compressing redundant visual information~\cite{shu2025videoxl,qin2025videoxl2,wang2025internvideo2,wang2025adaretake}.

For specific video understanding tasks, specialized architectures have been developed. Temporal action localization methods detect action boundaries using proposal networks or boundary regression~\cite{shi2023tridet}. Video grounding approaches align language queries with temporal segments through cross-modal matching~\cite{wu2025numberit,li2025llava-st}. Video question answering systems range from end-to-end VLMs to modular pipelines combining retrieval and reasoning components~\cite{li2024llava-onevision,zhang2025videollama3,li2024videochat-flash}.

\subsection{Agentic Framework}

The reasoning and action (ReAct) framework established the pattern of interleaving and action for task decomposition, enabling language models to interact with external tools and databases through iterative loops~\cite{yao2022react,shinn2023reflexion}. This paradigm has achieved significant development in the field of coding agent~\cite{yang2024swe-agent}, deep search~\cite{alzubi2025opendeepsearch} and deep research~\cite{li2025webthinker}.

Vision-language agents adapts these ideas to visual domains. Recent systems~\cite{su2025pixelreasoner,zheng2025deepeyes,pan2025dino-r1} focused on image understanding, defining action spaces including cropping, detection, and region-based question answering. These approaches demonstrated benefits in interpretability and error analysis through explicit action traces. Video Agents extend agentic reasoning to temporal understanding. Several architectural patterns have emerged: database interaction agents~\cite{luo2024videorag,tan2025rag-adapter} treat videos as queryable collections of clips or frames, using retrieval actions to gather relevant content. Tool-augmented agents~\cite{zhou2025reagent-v,yuan2025videodeepresearch} access specialized computer vision models for tasks like tracking, segmentation, or OCR.

Different training and design paradigms exist for video agents. Prompt-based agents~\cite{zhang2025dvd} rely on carefully designed prompts with powerful language models such as GPT-4 to guide decision-making. Reinforcement learning approaches~\cite{zhang2025vital,long2025m3-agent,zhou2025reagent-v} train agents through task-specific rewards to optimize action selection. Recent work has explored open-source agent implementations~\cite{yuan2025videodeepresearch} to address reproducibility concerns and reduce API costs. These efforts investigate whether smaller open-source models can match proprietary systems through better prompting, tool and architectural design.
\section{Method}
\label{sec:method}

In this section, we present the technical details of our proposed \textbf{A}gentic \textbf{V}ideo \textbf{I}ntelligence (AVI) framework. We first provide a formal problem definition and an overview of our approach. Then, we describe the offline construction of the structured video database. Finally, we elaborate on the agentic core architecture, including tool library design, the novel three-phase Retrieve-Perceive-Review reasoning process, and the agent runtime logic. The overall architecture of AVI is illustrated in ~\cref{fig_framework}.

\subsection{Overview and Problem Formulation}

Given a video $\mathcal{V}$ consisting of $T$ frames and a natural language question $q$, our goal is to generate an accurate answer $a$ through interpretable reasoning steps. Unlike single-pass VLM approaches that directly map $(\mathcal{V}, q) \rightarrow a$, we formulate video understanding as a sequential decision-making problem where an agent iteratively interacts with a structured environment $E$ to gather evidence before producing the final answer.

Formally, we model this as a Markov Decision Process (MDP) tuple $\mathcal{M} = (\mathcal{S}, \mathcal{A}, \mathcal{T}, \mathcal{W})$, where $\mathcal{S}$ represents the agent state containing conversation history, tool observations, and reasoning context. $\mathcal{A}$ denotes the action space comprising retrieval, perception, and review actions. $\mathcal{T}: \mathcal{S} \times \mathcal{A} \rightarrow \mathcal{S}$ defines state transitions through different tools. $\mathcal{W}$ represents the reflection signal of answer review.

\subsection{Structural Database Construction}
\label{subsec:database}

\subsubsection{Clip Segmentation and Captioning.} We first segment the video into a set of short clips $\{c_1, c_2, ..., c_N\}$ at intervals of $\Delta_t$. For each clip $c_i$ containing frames $f_i$, we generate a textual description $d_i$ to capture temporal narrative and event sequences using a lightweight vision-language model Qwen3-VL-8B~\cite{qwen3_vl_8b_hf}:
\begin{equation}
    d_i = \text{VLM}_{\text{caption}}(f_i).
\end{equation}

\subsubsection{Embedding Generation.} To enable semantic search, we compute dense embeddings for all captions using a text encoder:
\begin{equation}
    \mathbf{v}_i = \text{Encoder}_{\text{emb}}(d_i) \in \mathbb{R}^{d}.
\end{equation}
These embeddings, along with clip captions and other video meta information, are indexed in a vector database for efficient global search during agent runtime.

\subsubsection{Entity-Centric Graph Construction.} 
While clip-level captions provide temporal narrative structure, they fail to capture entity persistence and evolving relationships across distant temporal segments. Complex questions about entity interactions, state changes, or causal chains require explicit modeling of these cross-clip dependencies. The entity-centric graph addresses this limitation by capturing rich relational dynamics that unfold throughout the video.

\noindent \textbf{Knowledge Graph Formulation.} Formally, we construct a directed temporal knowledge graph $\mathcal{G} = (\mathcal{N}, \mathcal{R}, t_\text{edge})$ where:
\begin{itemize}
    \item $\mathcal{N} = \{n_1, n_2, ..., n_K\}$ represents entity nodes with their name, attributes, actions, and appearance, extracted from video captions
    \item $\mathcal{R} \subseteq \mathcal{N} \times \mathcal{N}$ denotes directed edges encoding relationships between entity nodes
    \item $t_\text{edge}$ maps each edge to temporal intervals $[t_{\text{start}}, t_{\text{end}}]$
\end{itemize}

\noindent \textbf{Graph Enrichment with Hierarchical Abstractions.} To facilitate multi-level reasoning, we augment the base entity graph with hierarchical abstractions. Super-nodes are created to represent entity groups, scenes, or events that span multiple atomic entities:
\begin{equation}
    n_{\text{super}} = \text{aggregate}(\{n_i : n_i \in \mathcal{C}_{\text{cluster}}\}),
\end{equation}
These abstractions enable the agent to reason at different semantic levels—from individual object interactions to scene-level events.

\noindent \textbf{Importance Weighting and Pruning.} Given that videos may contain numerous transient entities and relationships, we implement an importance scoring mechanism:
\begin{equation}
    w(n_i) = \lambda_1 \cdot \text{freq}(n_i) + \lambda_2 \cdot \text{centrality}(n_i) + \lambda_3 \cdot \text{query\_rel}(n_i, q),
\end{equation}
where frequency measures temporal persistence, centrality captures structural importance in the graph, and query relevance is computed dynamically based on the current question. This scoring enables selective graph traversal during inference, improving both efficiency and accuracy.

The entity-centric graph thus provides a structured substrate that preserves temporal dynamics, entity relationships, and hierarchical semantics—enabling sophisticated reasoning that would be intractable with flat video representations. This graph serves as the primary knowledge base for retrieve tools, supporting complex queries about entity interactions, temporal evolution, and causal relationships that are central to video understanding.

\subsection{Agentic Core Architecture}
\label{subsec:agent_core}

Our agent employs a three-phase reasoning protocol—Retrieve, Perceive, and Review—that mirrors human cognitive processes for video understanding. We also define two categories of executable tools that the agent can invoke in Retrieve and Perceive phases, respectively. In each of the three phase, the instruction prompt is distinguished to enhance the robustness of whole system. All the prompts are provided in the Appendix. 

\subsubsection{Retrieve Phase}
\label{subsubsec:retrieve}

\noindent \textbf{Retrieve Tools ($\mathcal{T}_r$).} 
The retrieve tool suite is designed to enable multi-granularity temporal exploration, addressing the fundamental challenge of efficiently navigating hours-long videos without exhaustive frame processing. Each tool serves a distinct and complementary role in the evidence gathering pipeline:

\begin{itemize}
\item \textit{clip\_retrieve}$(q_{\text{text}}, k) \rightarrow \{(c_i, s_i)\}_{i=1}^k$: This tool performs semantic similarity search over the clip embedding space, returning the top-$k$ most relevant clips ranked by cosine similarity $s_i = \cos(\mathbf{v}_i, \text{Encoder}_\text{emb}(q_{\text{text}}))$. This tool provides temporal anchors for subsequent fine-grained analysis.

\item \textit{clip\_merge}$(\{c_i\}_{i=1}^k) \rightarrow \{c_j\}_{j=1}^{k_m}$: Given that automatic clip segmentation may fragment continuous events across multiple segments, this tool intelligently merges adjacent or temporally proximate clips that share semantic coherence. This consolidation is essential for questions requiring understanding of extended sequences, such as temporal grounding.
    
\item \textit{global\_explore}$(q_{\text{text}}) \rightarrow G_{\text{summary}}$: This tool provides a query-focused traversal of the entire video timeline, generating a comprehensive summary and highlights potential relevant segments. This is particularly valuable when the agent needs to establish broader context before focusing on specific segments.
    
\item \textit{graph\_retrieve}$(e_{\text{query}}) \rightarrow \{(e_i, r_{ij}, t_{ij})\}$: Leveraging the entity-centric graph structure, this tool enables relational reasoning by retrieving entities and their interactions relevant to the query through graph traversal algorithms. This tool is indispensable when simple caption matching proves insufficient.
\end{itemize}

\noindent \textbf{Reasoning with Retrieve Tools.} 
Based on the above retrieve tools, the agent explores the video globally to identify candidate segments:
\begin{equation}
    \pi_r(\mathcal{A}_t^r | \mathcal{S}_t) = \text{LLM}(\mathcal{S}_t, \mathcal{T}_r, \text{prompt}_r),
\end{equation}
where $\text{prompt}_r$ instructs the agent to locate relevant timestamps without drawing conclusions yet in this phase. $\mathcal{A}_t^r=\{\mathcal{T}_r,\text{switch}_r\}$ contains retrieve tool execution along with phase switched from retrieve to perceive. Once the context is sufficient, the agent will response with no tool calls and generate a phase switch signal.

\subsubsection{Perceive Phase}

\noindent \textbf{Perceive Tools ($\mathcal{T}_p$).} 
While retrieve tools provide temporal localization based on textual abstractions, perceive tools perform direct visual analysis to extract precise, grounded information. Each perceive tool addresses specific visual understanding requirements:

\begin{itemize}
\item \textit{object\_detect}$(t_{\text{range}}, q_{\text{obj}}) \rightarrow \{(b_j, l_j, p_j)\}$: Employing Grounding-DINO~\cite{liu2024grounding-dino}, this tool performs open-vocabulary object detection within specified temporal ranges. For each detection, it returns bounding box coordinates $b_j \in \mathbb{R}^4$, class label $l_j$, and confidence score $p_j \in [0,1]$. This tool is essential for questions requiring precise object localization ("Where is X?"), counting ("How many X appear?"), or spatial relationship verification ("Is X near Y?").
    
\item \textit{text\_extract}$(t_{\text{range}}) \rightarrow \{text_\text{ocr}\}$: Utilizing an OCR model~\cite{cui2025paddleocr}, this tool extracts textual content from video frames, returning recognized text strings $text_\text{ocr}$.
    
\item \textit{boundary\_detect}$(t_{\text{range}}, q_{\text{event}}) \rightarrow [t_{\text{start}}, t_{\text{end}}]$: This tool performs fine-grained temporal localization by computing frame-level similarity scores using a CLIP model~\cite{tschannen2025siglip}. This tool is particularly valuable for action localization, scene transition detection, and temporal grounding tasks.
    
\item \textit{frame\_analysis}$(t_{\text{range}}, q_{\text{specific}}) \rightarrow a_{\text{analysis}}$: For complex visual reasoning that transcends the capabilities of specialized tools that use base CV models, this tool deploys a high-capacity Vision-Language Model (Qwen3-VL-8B~\cite{qwen3_vl_8b_hf}) to perform comprehensive frame analysis. While computationally expensive, this tool serves as a critical fallback and is used if the agent determines that lighter-weight tools are insufficient.
\end{itemize}

\noindent \textbf{Reasoning with Perceive Tools.}
The perceive tools suite embodies a carefully calibrated trade-off between computational cost and analytical capability. Within retrieved segments, the agent therefore can apply perceive tools for visual confirmation:
\begin{equation}
    \pi_p(\mathcal{A}_t^p | \mathcal{S}_t) = \text{LLM}(\mathcal{S}_t, \mathcal{T}_p, \text{prompt}_p).
\end{equation}
The perceive phase enforces visual grounding before answer generation. Similarly, $\mathcal{A}_t^p=\{\mathcal{T}_p,\text{switch}_p\}$ contains perceive tool execution along with phase switching from perceive to review. Once the evidence is sufficient, the agent will response with no tool calls and generate a phase switch signal.

\subsubsection{Review Phase}
In this phase, the agent evaluates all gathered context and evidence by itself with $\pi_v(\mathcal{S}_t)$  and decides whether to output the final answer or continue exploration:
\begin{equation}
    \mathcal{S}_{t+1} = \begin{cases}
        \text{output}(\mathcal{S}_t) & \text{if}\ \pi_v(\mathcal{S}_t)=True \\
        \text{re-perceive} & \text{otherwise}.
    \end{cases}
\end{equation}

\subsection{Agent Runtime}
\label{subsec:runtime}

\begin{algorithm}[t]
\caption{AVI Agent Execution}
\label{alg:avi_execution}
\begin{algorithmic}[1]
\Require Video database $\mathcal{K}$, Question $q$, Max iterations $N$
\Ensure Answer $a$
\State Initialize state $\mathcal{S}_0 \leftarrow \{\text{messages}: [q], \text{phase}: \text{retrieve}\}$
\State $t \leftarrow 0$
\While{$t < N$ and not terminated}
    \If{$\mathcal{S}_t.\text{phase} = \text{retrieve}$}
        \State $\mathcal{A}_t^r \sim \pi_r(\cdot | \mathcal{S}_t)$ \Comment{Select retrieval action}
        \If{$\mathcal{A}_t^r \in  \mathcal{T}_r$}
            \State $\mathcal{O}_t^r \leftarrow \text{Execute}(\mathcal{A}_t^r, \mathcal{K})$ \Comment{Query database}
        \Else
            \State $\mathcal{S}_t.\text{phase} \leftarrow \text{perceive}$ \Comment{Switch phase}
        \EndIf
    \ElsIf{$\mathcal{S}_t.\text{phase} = \text{perceive}$}
        \State $\mathcal{A}_t^p \sim \pi_p(\cdot | \mathcal{S}_t)$ \Comment{Select perception action}
        \If{$\mathcal{A}_t^p \in  \mathcal{T}_p$}
            \State $\mathcal{O}_t^p \leftarrow \text{Execute}(\mathcal{A}_t^p, \mathcal{V})$ \Comment{Process frames}
        \Else
            \State $\mathcal{S}_t.\text{phase} \leftarrow \text{review}$ \Comment{Switch phase}
        \EndIf
    \State $\mathcal{S}_{t+1} \leftarrow \text{UpdateState}(\mathcal{S}_t, \mathcal{A}_t, \mathcal{O}_t)$
    \State $t \leftarrow t + 1$
    \EndIf
    \If{$\pi_v(\mathcal{S}_t) = True$}
        \State \Return Output($\mathcal{S}_t$) \Comment{Output final answer}
    \Else
    \State $\mathcal{S}_t.\text{phase} \leftarrow \text{perceive}$ \Comment{Re-perceive}
    \EndIf
\EndWhile
\State \Return ForcedAnswer($\mathcal{S}_t$) \Comment{Fallback}
\end{algorithmic}
\end{algorithm}

\begin{table*}[ht]
\begin{center}
\caption{Comparison with state-of-the-art methods on video understanding benchmarks. We evaluate accuracy (\%) on LVBench, VideoMME-Long (without subtitles), and LongVideoBench, as well as temporal grounding performance (mIoU) on Charades-STA. Bold indicates best results. We divide all methods by whether is agentic system or trained by RL.}
\begin{tabular}{lccccccc}
\toprule
\textbf{Methods} & \textbf{Agentic} &\textbf{Trained}& \textbf{LVBench} &{\textbf{VideoMME}}&\multicolumn{2}{c}{\textbf{LongVideoBench}}&\textbf{Charades-STA}\\
&&&Overall&Long(w/o sub.)&Overall&Long&(mIoU)\\
\midrule
\textit{\textbf{Proprietary Methods}}\\
% \addlinespace[0.1cm]
GPT4-o\cite{hurst2024gpt4o}&No&Yes&48.9&65.3&66.7&60.9&35.7 \\
Gemini-1.5-Pro\cite{team2024gemini}&No&Yes&33.1&\textbf{67.4}&64.0&58.6&25.0 \\
OpenAI o3\cite{openaio3}&No&Yes&57.1&64.7&67.5&60.6&- \\
\midrule
\textit{\textbf{Open-Source Methods}}\\
% \addlinespace[0.1cm]
mPLUG-Owl3\cite{ye2024mplug}&No&Yes&43.5&50.1&59.8&-&- \\
TimeSuite\cite{zeng2024timesuite}&No&Yes&-&41.9&59.2&-&57.5 \\
VideoChat-Flash\cite{li2024videochat-flash}&No&Yes&42.9&44.9&66.5&-&- \\
Qwen2.5-VL-72B\cite{bai2025qwen2.5vl}&No&Yes&47.7&63.9&60.7&-&50.9 \\
\midrule
\textit{\textbf{Agentic Frameworks}}\\
% \addlinespace[0.1cm]
VideoAgent\cite{wang2024videoagent}&Yes&No&29.3&49.0&-&-&- \\
VideoTree\cite{wang2025videotree}&Yes&No&28.8&54.2&-&-&-\\
ReAgent-V\cite{zhou2025reagent-v}&Yes&Yes&40.5&49.8&-&-&-\\
VITAL\cite{zhang2025vital}&Yes&Yes&-&54.0&-&-&59.9\\
StreamAgent\cite{yang2025streamagent}&Yes&No&-&50.6&57.9&-&-\\
ViTL\cite{wang2025ViTL}&Yes&Yes&47.4&52.7&63.3&-&54.0\\
\midrule
AVI(ours)&Yes&No&\textbf{61.4}&59.8&\textbf{68.4}&\textbf{62.8}&\textbf{60.0}\\
\bottomrule
\end{tabular}
% \vspace{-1em}
\label{table1}
\end{center}
\end{table*}

~\cref{alg:avi_execution} presents the runtime execution flow. The agent maintains a state $\mathcal{S}_t$ containing conversation history $\mathcal{H}_t$, tool observations $O_t$, and the current reasoning phase. At each timestep, the agent: \textbf{i) Thought Generation:} Analyzes the current state and plans next action through chain-of-thought reasoning. \textbf{ii) Action Selection:} Chooses a tool from the available action space based on the current phase. \textbf{iii) Observation Integration:} Executes the tool and incorporates results into the state. \textbf{iv) Phase Transition:} Evaluates progress and potentially transitions between phases.

The execution continues until either the agent generates a confident answer in the review phase, or the maximum iteration limit is reached. To ensure stable convergence, we implement several mechanisms:

\paragraph{Tool Call Constraints.} Each phase restricts available tools to prevent action space explosion:
\begin{equation}
    \mathcal{A}_t = \begin{cases}
        \mathcal{T}_r & \text{if phase} = \text{retrieve} \\
        \mathcal{T}_p & \text{if phase} = \text{perceive} \\
        \{\text{output}\} & \text{if phase} = \text{review}
    \end{cases}
\end{equation}

\paragraph{Open-Source Model Ensemble.} Unlike prior work requiring proprietary APIs, we leverage open-source models: Qwen3-32B~\cite{yang2025qwen3} for action planning, thought and observation; CLIP~\cite{tschannen2025siglip} for boundary detection, Grounding-DINO~\cite{liu2024grounding-dino} for detection, PaddleOCR~\cite{cui2025paddleocr} for text extraction; Qwen3-VL-8B~\cite{qwen3_vl_8b_hf} for frame analysis when needed.

This ensemble achieves competitive performance while maintaining reproducibility and cost-efficiency. The total computational cost per question is:
\begin{equation}
    \text{Cost} = \underbrace{C_{\text{db}}}_{\text{one-time}} + \underbrace{N \cdot C_{\text{LLM}}}_{\text{reasoning}} + \underbrace{\sum_{i=1}^{N_{\text{tools}}} C_{\text{tool}_i}}_{\text{tool action}}
\end{equation}
where database construction cost $C_{\text{db}}$ is amortized across multiple questions on the same video.
\section{Experiments}
\subsection{Experimental Setups}
\paragraph{Datasets.} We evaluate AVI on four challenging video understanding benchmarks: \textbf{LVBench}~\cite{wang2025lvbench} contains 1,549 questions across 103 web videos averaging 68.4 minutes, covering six task categories: Event Recognition (ER), Event Understanding (EU), Key Information Retrieval (KIR), Temporal Grounding (TG), Reasoning (Rea), and Summarization (Sum).  \textbf{VideoMME-Long}~\cite{fu2025videomme} contains 900 questions and 300 extra-long videos with durations ranging from 30 minutes to 1 hour. \textbf{LongVideoBench}~\cite{wu2024longvideobench} on the validation set comprises 1337 videos and its long subset comprises 566 questions on 188 videos with durations ranging from 15 minutes to 1 hour. \textbf{Charades-STA}~\cite{gao2017charades-sta} evaluates temporal grounding capabilities with 3,720 language queries mapped to temporal segments in 1336 household activity videos, measured by mean Intersection over Union (mIoU).

\paragraph{Implementary Details.} The reasoning agent employs Qwen3-32B-Instruct~\cite{yang2025qwen3} as the core orchestration LLM due to its agentic ability, with a maximum of 10 iterations per question. All the embedding models employ Qwen3-Embedding-4B~\cite{zhang2025qwen3-emb}. For database construction, we segment videos into 5-second clips and generate captions using Qwen3-VL-8B~\cite{qwen3_vl_8b_hf}. Entity graphs are extracted also using Qwen3-32B-Instruct. The raw frames are sampled from the videos with $\text{fps}=2$. All frames are resized to 720p to maintain visual details. In each tool calling, most of the arguments are determained by the core LLM itself. Some manually set specific tool arguments are: $\text{top-}k=16$ in \textit{clip\_retrieve}, $\text{iou\_threshold}=0.5$ in \textit{object\_detect}, $\text{max\_frames}=64$ in \textit{frame\_analysis}. For evaluation, we instruct the agent to construct the answer as a specific format and then extract the results by regular expression matching. All experiments use 8×A800 GPUs for inference. 

\subsection{Comparison to state-of-the-arts}

We evaluate AVI against proprietary VLMs, open-source VLMs, and other agent-based methods.\cref{table1} shows that AVI achieves 61.4\% on LVBench, 59.8\% on VideoMME-Long, and 62.8\% on LongVideoBench-Long. It achieves competitive performance on these benchmarks. 

Specifically, on LVBench, AVI surpasses OpenAI o3 by 4.3\% and Qwen2.5-VL-72B by 13.7\%, suggesting that decomposing video understanding into structured sub-tasks can be more effective than processing thousands of visual tokens at once. The performance gap is even more pronounced when compared to other agent-based methods—AVI nearly significantly exceed the performance of VideoAgent, VideoTree and VITAL on VideoMME-Long despite VITAL trains the model with agentic RL. Notably, AVI also achieves 60.0\% on Charades-STA. This flexibility to adapt to diverse task formats—from multiple-choice QA to exact temporal localization—highlights a key advantage of our modular design.

\begin{table}[!t]
\begin{center}
\small  % 或使用 \footnotesize 来调整字体大小
\setlength{\tabcolsep}{3pt}  % 调整列间距，默认是6pt
\caption{Task-specific performance breakdown on LVBench: Event Recognition (ER), Event Understanding (EU), Key Information Retrieval (KIR), Temporal Grounding (TG), Reasoning (Rea), and Summarization (Sum).}
\begin{tabular}{lccccccc}
\toprule
\textbf{Methods} & \textbf{ER} & \textbf{EU} &\textbf{KIR}&\textbf{TG}&\textbf{Rea}&\textbf{Sum}&\textbf{Avg}\\
\midrule
% \addlinespace[0.1cm]
GPT4-o\cite{hurst2024gpt4o}&48.9&49.5&48.1&40.9&50.3&50.0&48.9 \\
Gemini-1.5-Pro\cite{team2024gemini}&32.1&30.9&39.3&31.8&27.0&32.8&33.1 \\
OpenAI o3\cite{openaio3}&57.6&56.4&62.9&46.8&50.8&\textbf{67.2}&57.1 \\
% \addlinespace[0.1cm]
Qwen2.5-VL-72B\cite{bai2025qwen2.5vl}&-&-&-&-&-&-&47.7 \\
InternVL2.5-78B\cite{chen2024internvl2.5}&43.8&42.0&42.1&36.8&51.0&37.9&43.6 \\
VideoChat-Flash\cite{li2024videochat-flash}&51.1&46.0&49.0&38.9&48.5&34.5&48.2 \\
% \addlinespace[0.1cm]
VideoAgent\cite{wang2024videoagent}&28.0&30.3&28.0&29.3&28.0&36.4&29.3 \\
VideoTree\cite{wang2025videotree}&30.3&25.1&26.5&27.7&31.9&25.5&28.8\\
AVI(ours)&\textbf{62.7}&\textbf{57.3}&\textbf{71.3}&\textbf{59.7}&\textbf{56.2}&66.0&\textbf{61.4}\\
\bottomrule
\end{tabular}
% \vspace{-1em}
\label{table2}
\end{center}
\end{table}

~\cref{table2} breaks down LVBench performance by task category. AVI excels particularly in Key Information Retrieval (71.3\%) and Summarization (66.0\%), demonstrating the effectiveness of our structured database for evidence gathering. The strong performance on Temporal Grounding (59.7\%) validates our three-phase reasoning design for temporal localization.

\subsection{Qualitative Analysis}

\begin{figure*}[!ht]
    \centering
    \includegraphics[width=\linewidth]{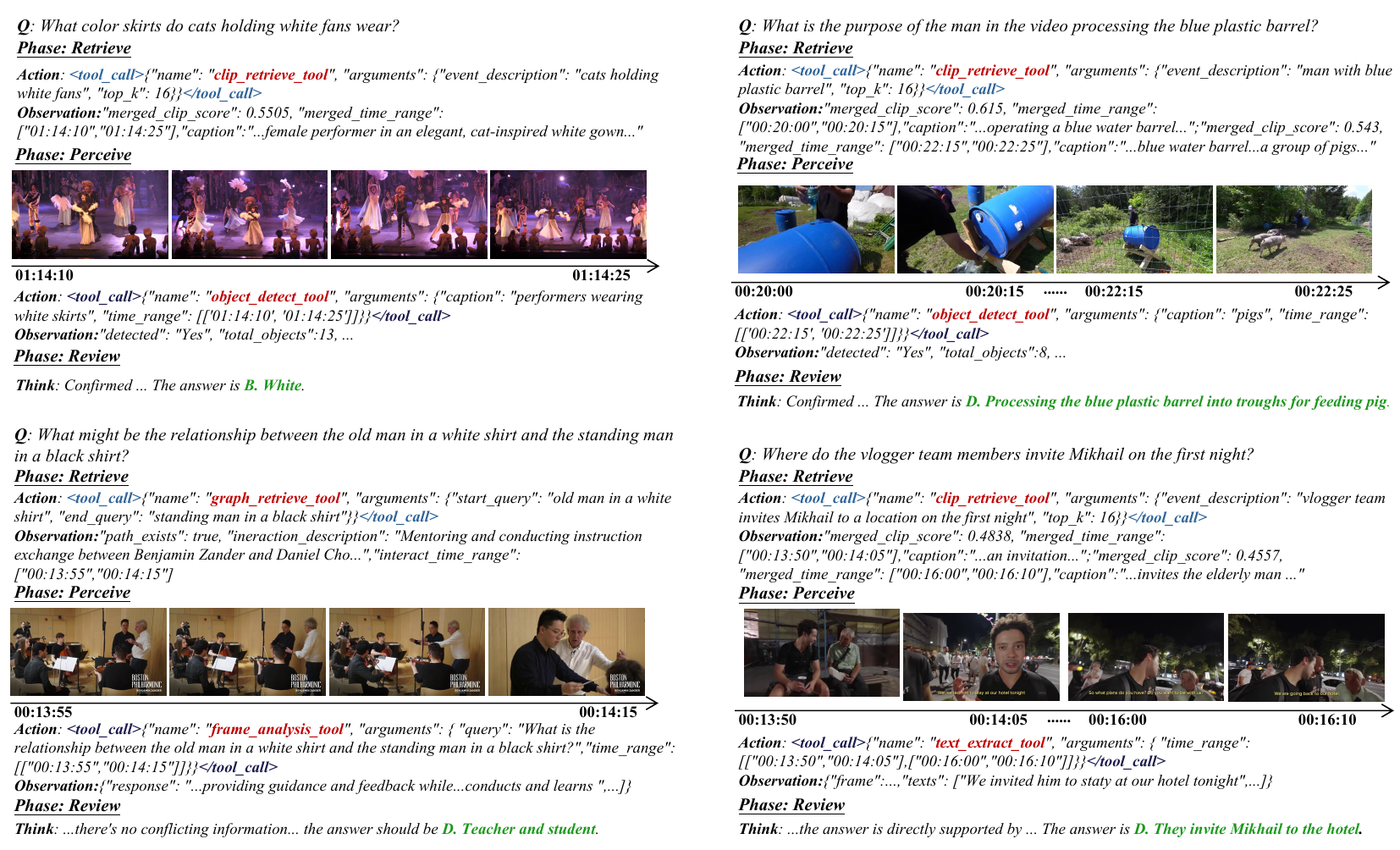}
    \caption{Qualitative examples of AVI's reasoning traces on complex video understanding tasks. The trace shows how AVI combines text-based retrieval with visual analysis to identify both the temporal segment and spatial semantics. Each phase is clearly delineated, demonstrating the interpretability of our approach.}
    \label{fig_cases}
\end{figure*}

~\cref{fig_cases} illustrates representative reasoning traces from AVI across different video understanding scenarios. For example, in the first case (top left), the agent successfully answers through systematic exploration: the Retrieve phase identifies relevant clips containing "cat performers" and "white skirts," the Perceive phase deploys \textit{object\_detect} tool to locate performers wearing white skirts, and the Review phase synthesizes observations to confirm the agent can answer without conflict. Other cases show our AVI's different capability such as relationship identification by \textit{graph\_retrieve} tool, OCR on specific frames by \textit{text\_extract} tool, showcasing the complementary nature of our multi-granularity tools. These traces highlight AVI's interpretability advantage—each decision and observation is explicitly recorded, enabling error diagnosis and trust calibration.

\subsection{Ablation Studies}

\begin{table}[!t]
\begin{center}
\small  % 或使用 \footnotesize 来调整字体大小
\setlength{\tabcolsep}{6pt}  % 调整列间距，默认是6pt
\caption{Ablation studies on LVBench and VideoMME-Long for the the key architectural components.}
\begin{tabular}{c|cc}
\toprule
\textbf{Methods} & \textbf{LVBench} & \textbf{VideoMME-Long}\\
\midrule
AVI w/o three-phase&60.4&58.2\\
AVI w/o entity graph&59.8&57.5\\
AVI w/o review phase&60.2&58.6\\
AVI&61.4&59.8\\
\bottomrule
\end{tabular}
% \vspace{-1em}
\label{table3}
\end{center}
\end{table}

~\cref{table3} validates our key architectural choices through systematic ablation: Removing the structured three-phase design and allowing free-form tool selection decreases performance on LVBench and VideoMME-long. Without phase constraints, agents often skip retrieval and directly attempt perception on random segments, missing crucial context. The performance also decreases when removing review phase due to a lack of self-reflection, evidence revisit and iterative refinement. The phase structure enforces sufficient exploration before detailed analysis. Ablating the entity graph while retaining clip-level retrieval reduces LVBench accuracy to 59.8\% and VideoMME to 57.5\%, confirming the graph's importance for modeling persistent entities and their interactions across temporal boundaries.

\subsection{Analysis on Agentic System}

\begin{figure}
    \centering
    \includegraphics[width=\linewidth]{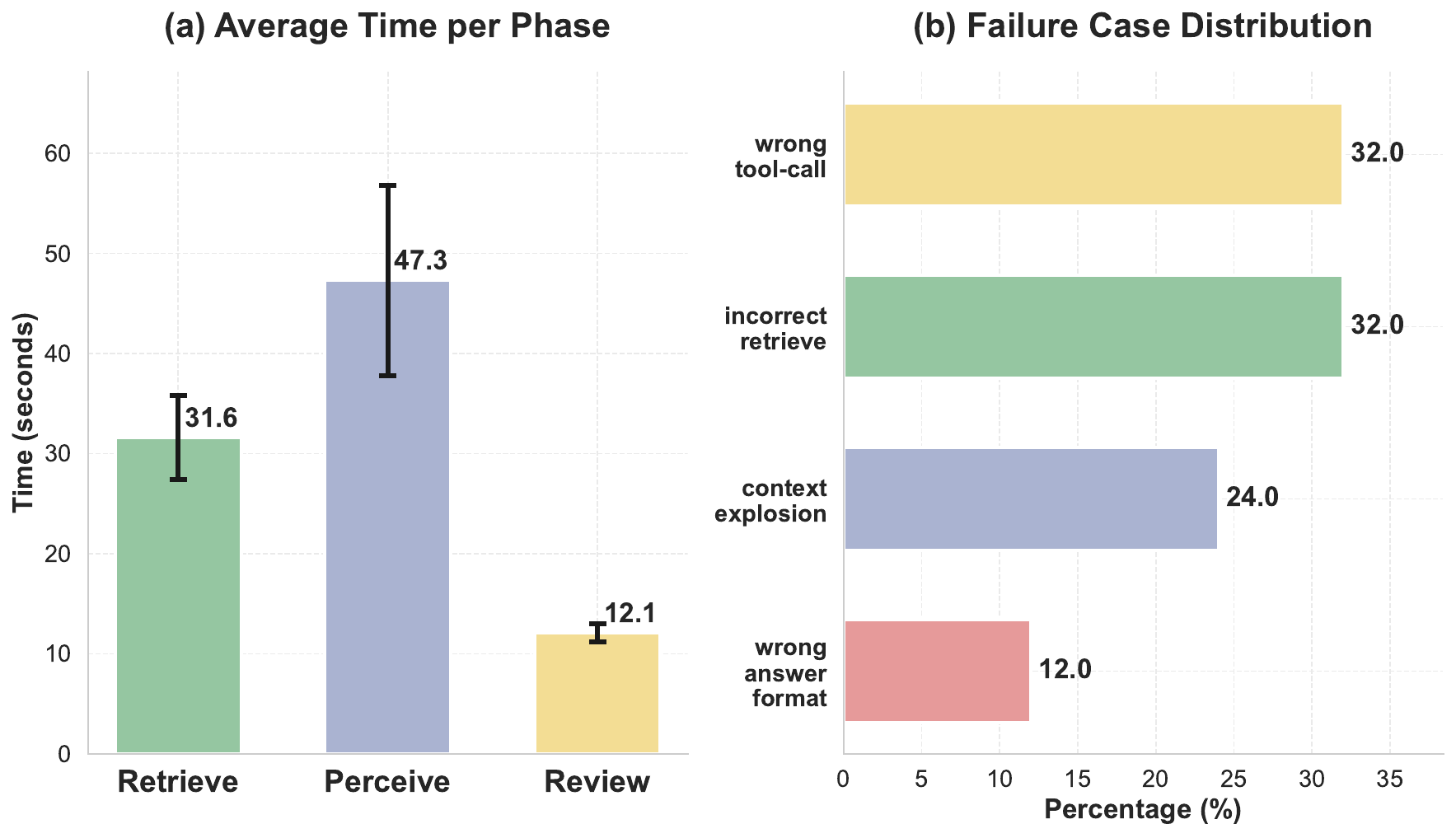}
    \caption{Analysis on the (a) average time per phase and (b) failure case distribution on LVBench.}
    \label{fig_bar}
    % \vspace{-1em}
\end{figure}

We evaluate AVI on videos of varying lengths: performance remains stable from 10-minute to 1-hour videos, while database construction scales linearly with video length. The average iteration of AVI in the cases of LVBench is 5.3. Unlike black-box VLMs, AVI's explicit traces enable systemic analysis and targeted improvements. As illustrated in ~\cref{fig_bar}, on LVBench, the left bar shows the time (avg and std) per phase and the right bar shows the distribution for 35 failure cases with different reasons. The most time-consuming phase is perceive phase, as shown in ~\cref{fig_bar}(a). It is because in perceive phase the raw frames need to be processed, which also indicates the significance of the three-phase design. In the failure case analysis, as shown in ~\cref{fig_bar}(b), most of  the cases fail due to wrong tool-call or incorrect retrieved context, which is limited by the agentic ability of the main LLM. Thus, in this work we further enhance AVI by adjusting the system prompt and  context engineering. Codes and prompts are in the appendix.
\section{Conclusion}

We presented Agentic Video Intelligence (AVI), a training-free agentic framework that achieves state-of-the-art performance on complex video understanding through systematic design rather than model parameter scaling or extensive training. By combining a structured multi-modal database, a three-phase reasoning mechanism (Retrieve-Perceive-Review), and diverse retrieve and perceive tools, AVI surpasses OpenAI o3 and also outperforms existing agentic systems on several benchmarks, while providing interpretable reasoning traces. Our work demonstrates that careful architectural design can match or exceed massive VLMs and RL-trained agents. AVI suggests a compositional, training-free approach that offers a promising path forward for video understanding, balancing performance, efficiency, and interpretability. Future work will explore dynamic database updates during inference and parallel execution strategies.
{
    \small
    \bibliographystyle{ieeenat_fullname}
    \bibliography{main}
}

\clearpage
\setcounter{page}{1}
\maketitlesupplementary

\section{Implementation Details}

\subsection{Clip-Level Caption Generation}

The prompt for the clip-level caption generation is illustrated in ~\cref{prompt_caption}.

\subsection{Entity-Centric Graph Generation}

The entity extraction process operates on clip-level captions to identify and track entities across the video. The prompt for initial graph generation is in ~\cref{prompt_graph}.

\subsubsection{Hierarchical Abstraction Construction}

Super-nodes are created through spectral clustering on the entity similarity matrix:
\begin{equation}
    \mathbf{S}_{ij} = \text{sim}(n_i, n_j) \cdot \mathbb{1}[\exists \text{ path}(n_i, n_j, k \leq 2)],
\end{equation}
where the indicator function ensures only connected components within 2-hop distance are considered for clustering.

The number of clusters $k$ is determined by the elbow method on the eigenvalue spectrum, typically ranging from 3 to 8 for hour-long videos. Each cluster with more than 3 entities forms a super-node, inheriting the union of temporal spans and the intersection of common attributes from its constituents.

\subsubsection{Importance Scoring Components}

The importance weight $w(n_i)$ for each node combines three factors with specific coefficients:

\noindent \paragraph{Frequency Score ($\lambda_1 = 0.4$):}
\begin{equation}
    \text{freq}(n_i) = \frac{|\{c_j : n_i \in c_j\}|}{|\mathcal{C}|} \cdot \log\left(1 + \sum_{j} \Delta t_{ij}\right),
\end{equation}
where $|\mathcal{C}|$ is the total number of clips and $\Delta t_{ij}$ is the duration of appearance $j$.

\noindent \paragraph{Centrality Score ($\lambda_2 = 0.3$):}
\begin{equation}
    \text{centrality}(n_i) = 0.5 \cdot \text{degree}(n_i) + 0.5 \cdot \text{between}(n_i),
\end{equation}
where degree centrality is normalized by maximum degree, and betweenness centrality captures the node's role in connecting different graph components.

\noindent  \paragraph{Query Relevance Score ($\lambda_3 = 0.3$):}
\begin{equation}
    \text{query\_rel}(n_i, q) = \text{cos\_sim}(\mathbf{v}_{n_i}, \mathbf{v}_q),
\end{equation}
where $\mathbf{v}_{n_i}$ and $\mathbf{v}_q$ are text embeddings of the node name and query respectively.

\subsection{Prompts in the Agentic Runtime.}

\begin{figure}[!t]
\begin{tcolorbox}
[
colframe=black!40!white,
colback=gray!10!white,
sharp corners=south,
title=\textbf{The whole system prompt of AVI.}
]
You are a Protocol-Driven Video Analysis Engine.

**Current Stage: {\textit{\textless current\_stage \textgreater}}**

{\textit{\textless core\_protocol \textgreater}~\cref{prompt_core_protocol}}

{\textit{\textless stage\_instruction \textgreater}~\cref{prompt_retrieve} or ~\cref{prompt_perceive} or ~\cref{prompt_review}}

Here are tools you can use:

{\textit{\textless tool\_descriptions \textgreater}}~\cref{tool_des}

Total video length: {\textit{\textless video\_length \textgreater}} seconds.

\end{tcolorbox}
\caption{The whole system prompt of AVI.}
\label{prompt_avi}
\vspace{-1em}
\end{figure}

The entire system prompt of the proposed AVI is in ~\cref{prompt_avi}. The core protocol for all three phases is in ~\ref{prompt_core_protocol} while the separate instruction for each phase is in ~\ref{prompt_retrieve}, ~\cref{prompt_perceive}, and ~\cref{prompt_review}, respectively.

Moreover, we append the detailed tool descriptions in the system prompt to enhance the AVI's comprehension of each rool's role and value, as shown in ~\cref{tool_des}.

\section{More Experimental Results.}
\subsection{More Results on LVBench and VideoMME-Long.}

\begin{table}[!h]
\begin{center}
\small  % 或使用 \footnotesize 来调整字体大小
\setlength{\tabcolsep}{3pt}  % 调整列间距，默认是6pt
\caption{Video task-specific performance breakdown on LVBench and VideoMME-Long.}
\begin{tabular}{lcc}
\toprule
\textbf{Video Type} & \textbf{Gemini-1.5-pro} & \textbf{AVI(ours)}\\
\midrule
\textbf{LVBench}\\
\textit{Sports}&41.2&66.2\\
\textit{Documentary}&25.4&73.6\\
\textit{Event Record}&28.2&72.2\\
\textit{Lifestyle}&36.0&65.5\\
\textit{TV Show}&36.4&51.6\\
\textit{Cartoon}&29.7&62.5\\
\textit{Overall}&33.1&61.4\\
\midrule
\textbf{VideoMME-Long}\\
\textit{Knowledge}&73.4&65.1\\
\textit{File\&Television}&70.1&49.2\\
\textit{Sports Competition}&58.3&60.1\\
\textit{Artistic Performance}&63.6&46.7\\
\textit{Life Record}&65.1&57.9\\
\textit{Multilingual}&70.8&60.0\\
\textit{Overall}&67.4&59.8\\
\bottomrule
\end{tabular}
% \vspace{-1em}
\label{table_video_type}
\end{center}
\end{table}

Table~\ref{table_video_type} presents fine-grained performance breakdown across diverse video categories. On LVBench, AVI demonstrates substantial improvements across all video types, with particularly striking gains in Documentary (+48.2\%) and Event Record (+44.0\%) categories. These categories demand understanding of complex temporal narratives and entity interactions—precisely the capabilities enabled by our structured environment and three-phase reasoning framework. The consistent superiority across all categories (28.3\% overall improvement) validates that our human-inspired approach—combining global exploration through entity graphs with local visual analysis—effectively handles diverse video content without task-specific adaptations.

On VideoMME-Long, while Gemini-1.5-pro maintains advantages in knowledge-intensive categories, AVI shows competitive performance in action-oriented domains like Sports Competition. This pattern aligns with our framework's design philosophy: rather than relying on massive parameter counts to encode world knowledge, AVI leverages structured retrieval and tool-integrated reasoning to understand video content. The results demonstrate that our open-source ensemble approach can achieve strong performance without proprietary APIs, significantly improving accessibility for the research community.

\subsection{More Results on Charades-STA.}

\begin{table}[!h]
\begin{center}
\small  % 或使用 \footnotesize 来调整字体大小
\caption{Comparison of methods on video temporal grounding benchmark Charades-STA.}
\begin{tabular}{lcccc}
\toprule
\textbf{Methods} & \textbf{R@0.3} & \textbf{R@0.5} &\textbf{R@0.7}&\textbf{mIoU}\\
\midrule
Qwen2.5-VL-72B\cite{bai2025qwen2.5vl}&-&-&-&50.9 \\
Qwen2.5-VL-7B\cite{bai2025qwen2.5vl}&67.9&50.3&24.3&43.6 \\
TimeSuite~\cite{zeng2024timesuite}&69.9&48.7&24.0&57.5\\
VideoChat-Flash\cite{li2024videochat-flash}&74.5&53.1&27.6&48.4 \\
VITAL\cite{zhang2025vital}&83.1&\textbf{72.0}&\textbf{46.7}&59.9\\
AVI(ours)&\textbf{88.6}&\textbf69.0&37.6&\textbf{60.0}\\
\bottomrule
\end{tabular}
% \vspace{-1em}
\label{table_charades-sta}
\end{center}
\end{table}

To evaluate AVI's capability beyond question-answering tasks, we conduct experiments on Charades-STA for temporal grounding. As shown in Table~\ref{table_charades-sta}, AVI achieves the highest R@0.3 score (88.6\%) and mIoU (60.0\%), surpassing previous best methods including the RL-trained VITAL. This strong performance validates a key insight: our Retrieve-Perceive-Review framework naturally supports temporal localization through iterative refinement. The Retrieve phase first identifies coarse temporal segments via entity graph traversal, while the Perceive phase zooms in for precise boundary detection—mirroring how humans locate specific moments in videos.

Notably, AVI achieves these results using only open-source models without any task-specific training, whereas competitors like VITAL require resource-intensive RL training. The competitive R@0.5 performance (69.0\%) further demonstrates that our structured environment and tool-integrated reasoning provide sufficient temporal understanding for practical applications. These results underscore AVI's versatility: a single framework that excels at both complex video understanding and fine-grained temporal grounding, while maintaining interpretability and accessibility through its human-inspired design.

\begin{figure}[!t]
\begin{tcolorbox}
[
colframe=black!40!white,
colback=gray!10!white,
sharp corners=south,
title=\textbf{The tool descriptions.}
]
\textit{'global\_explore\_tool'}: '• To get a global information about events and main subjects in the video, use `global\_explore\_tool`.',

\textit{'clip\_retrieve\_tool'}: '• To retrieve without a specific timestamp, use `clip\_retrieve\_tool`.',

\textit{'graph\_retrieve\_tool'}: '• To retrieve for **relationships or paths** between two entities (subjects) or events, use `graph\_retrieve\_tool`.',     

\textit{'frame\_analysis\_tool'}: '• If the retrieved material lacks precise, question-relevant detail (e.g., an unknown name), call `frame\_analysis\_tool` with a list of time ranges.',

\textit{'object\_detect\_tool'}: '• To perform open-set object detection on video frames, use `object\_detect\_tool` with time ranges and text description of objects to detect.',

\textit{'boundary\_detect\_tool'}: '• To detect event boundaries (start/end points) in video clips, use `boundary\_detect\_tool` with event description and time ranges.',

\textit{'text\_extract\_tool'}: '• To performs text recognition on video frames, use `text\_extract\_tool` with time ranges.'

\end{tcolorbox}
\caption{The tool descriptions.}
\label{tool_des}
\end{figure}

\begin{figure*}[!ht]
\begin{tcolorbox}
[
colframe=black!40!white,
colback=gray!10!white,
sharp corners=south,
title=\textbf{The Prompt for Caption Generation}
]
You are an expert video analysis assistant. Your task is to generate detailed, objective, and accurate descriptions for video clips. These descriptions will be used to build a database and environment for an AI video agent, so precision and comprehensiveness are crucial.

You will receive a sequence of consecutive frames from a video clip. Please thoroughly understand the content of this clip and output a JSON object strictly adhering to the template provided below.

\vspace{1em}

\textbf{Description Requirements:}

\quad \textbf{1.  Objectivity}: Describe only what is **actually visible** in the video. Avoid any speculation, interpretation, or subjective judgment.

\quad \textbf{2.  Detail}:

    \quad\quad •   **Subjects \& Objects:** Identify all significant subjects (people, animals) and objects (items, vehicles, etc.) present.
    
    \quad \quad•   **Attributes:** Detail key attributes of these subjects and objects (e.g., color, size, state, position).
    
    \quad \quad•   **Actions \& Behaviors:** Describe the actions and behaviors performed by subjects.
    
    \quad \quad•   **Interactions:** If applicable, describe interactions between subjects, or between subjects and objects.
    
    \quad \quad•   **Scene \& Environment:** Describe the background, environment, and any changes in the scene.
    
\quad  \textbf{3. Temporal Order}: The narration must strictly follow the **chronological order** of events as they unfold in the video clip.

\quad \textbf{4.  Smoothness}: Use natural, flowing language, as if providing a voice-over narration.

\quad \textbf{5.  No Timestamps in Description}: The `clip\_description` content should **not** include `clip\_start\_time` or `clip\_end\_time`, as these are provided separately in the JSON structure.

\vspace{1em}

\textbf{Output Template}:

\{\\
\quad  "\textit{clip\_start\_time}": CLIP\_START\_TIME\_IN\_SECONDS,\\
\quad  "\textit{clip\_end\_time}": CLIP\_END\_TIME\_IN\_SECONDS,\\
\quad "\textit{subject\_registry}": \{

    \quad \textit{subject\_i}: \{
    
      \quad "\textit{name}": fill with short identity if name is unknown,
      
      \quad "\textit{appearance}": list of appearance descriptions,
      
      \quad "\textit{identity}": list of identity descriptions,
      
      \quad "\textit{first\_seen}": timestamp\},...\},
      
  \quad"\textit{clip\_description}": "A smooth, detailed, objective, and chronologically ordered natural language narration of the video clip content, including all significant subjects, objects, actions, interactions, and scene changes"\\
\}
\end{tcolorbox}
\caption{The Prompt for Caption Construction.}
\label{prompt_caption}
\end{figure*}

\begin{figure*}[!ht]
\begin{tcolorbox}
[
colframe=black!40!white,
colback=gray!10!white,
sharp corners=south,
title=\textbf{The Prompt for Graph Generation.}
]
You are a professional Video Content Analyst.

Your task is to analyze a complete, chronological list of video captions provided by the user.

\vspace{1em}

\textbf{Core Objectives:}
    
    \quad 1.  **Analyze Human Subjects:** 
        
        \quad\quad • Focus **exclusively on human subjects**.
        
        \quad\quad • If a human has a pre-defined ID (e.g., 'Subject\_100', 'Person\_12') mentioned in the caption, you **MUST** place this ID in the `subject\_id` field.
        
        \quad\quad • You **MUST** also provide a descriptive `subject\_name` (e.g., 'Man in red shirt', 'Anna').
        
        \quad\quad • If no pre-defined ID is present, the `subject\_id` field **MUST** be `null`.
        
        \quad\quad • Consolidate each subject's attributes, *total* appearance timeline, and their *individual* actions (e.g., 'walks across room', 'sits down').
        
    \quad 2.  **Analyze Interactions:** 
    
        \quad \quad • Identify all **INTERACTIONS** between two or more identified human subjects.
    
        \quad\quad • Log these events in the top-level `interactions` list.
        
        \quad\quad • **CRITICAL:** When populating the `subjects\_involved` list for an interaction, you **MUST** use the `subject\_id` (e.g., 'Subject\_100') of the subjects, not their descriptive `subject\_name`.

\vspace{1em}

\textbf{JSON Schema (Strict):}

The root output **MUST** be a JSON object with two keys: `video\_analysis` and `interactions`.

    ```json\\
    \{\\
          "\textit{subject\_id}": "Subject\_100", // (The ID, e.g., 'Subject\_100', or null if none)\\
          "\textit{subject\_name}": "Man in red shirt", // (The descriptive name)\\
          "\textit{appearance\_timeline}": [["start\_time\_str", "end\_time\_str"]],\\
          "\textit{attributes}": ["attr 1", "attr 2"],
          
           "\textit{actions\_events}": [\{
           
              \quad "\textit{action}": "The specific *individual* action performed (e.g., 'sits down')",
              
              \quad   "\textit{timestamp}": ["start\_time\_of\_action", "end\_time\_of\_action"]\}]\}],\\
      "interactions": [\{
      
          \quad "\textit{subjects\_involved}": ["Subject\_100", "Subject\_101"], // (Must use subject\_id)
          
          \quad "\textit{interaction\_description}": "A clear description of the interaction (e.g., 'Subject\_100 gives book to Subject\_101')",
          
          \quad "\textit{timestamp}": ["start\_time\_of\_interaction", "end\_time\_of\_interaction"]\\
        \}\\
    ```
\end{tcolorbox}
\caption{The Prompt for Graph Construction.}
\label{prompt_graph}
\end{figure*}

\begin{figure*}[!t]
\begin{tcolorbox}
[
colframe=black!40!white,
colback=gray!10!white,
sharp corners=south,
title=\textbf{Core protocols for AVI.}
]
\textbf{\textless CORE\_EXECUTION\_LOOP \textgreater}

Follow the THINK → ACT → OBSERVE loop strictly:

\quad • THOUGHT: Reason step-by-step about the current state and plan the next action.

\quad • ACTION: Call exactly one function.

\quad • OBSERVATION: Summarize the function's output.

\textbf{\textless CORE\_EXECUTION\_LOOP \textgreater}

\vspace{1em}

\textbf{\textless CORE\_PROTOCOL \textgreater}

\quad 1.  **Tool Call Limit:** In each iteration, you MUST call **EXACTLY ONE** tool. NEVER call multiple tools in a single response.

\quad 2.  **Argument Integrity:** Only pass arguments that come verbatim from the user or from earlier function outputs. **NEVER invent them**.

\quad 3.  **Data Integrity (Caption Reliance - GLOBAL RULE):** The captions provided by retrieve tools are **UNRELIABLE HINTS** for location only. You are **ABSOLUTELY FORBIDDEN** from drawing final conclusions or answering the question based solely on text captions. Final answers require visual confirmation in the Perception Stage.

\textbf{\textless CORE\_PROTOCOL \textgreater}
\end{tcolorbox}
\caption{Core protocols for AVI.}
\label{prompt_core_protocol}
\end{figure*}

\begin{figure*}[!t]
\begin{tcolorbox}
[
colframe=black!40!white,
colback=gray!10!white,
sharp corners=south,
title=\textbf{Instruction for Retrieve Phase.}
]
\textbf{\textless RETRIEVE\_STAGE\_INSTRUCTION \textgreater}

**\textbf{Current Goal (STRICTLY LIMITED)}:** The **SOLE** purpose of this stage is to identify and locate the most promising time segments (timestamps) for later visual confirmation.

\vspace{1em}

\textbf{\textless RETRIEVE\_STAGE\_PROHIBITIONS \textgreater}

\quad 1.  **NEVER ANSWER:** You are **ABSOLUTELY FORBIDDEN** from providing the final answer in this stage, regardless of how conclusive the captions seem. Answering now is a protocol failure.

\quad 2.  **Caption Reliability:** Treat captions as navigational aids, not facts. Their primary value is providing time coordinates.

\textbf{\textless RETRIEVE\_STAGE\_PROHIBITIONS \textgreater}

\vspace{1em}

**\textbf{Tool Usage Strategy}:**

\quad 1.  **Primary Retrieve:** Start with `clip\_retrieve\_tool`.

\quad 2.  **Broad Context:** If `clip\_retrieve\_tool`'s output is insufficient, inaccurate, or too limited, then call `global\_explore\_tool` for a high-level summary.

\vspace{1em}

**\textbf{Stage Completion \& Switch}:**

Once you have located the relevant time-spans or textual information necessary to proceed to visual analysis (Perception Stage), you must include the following explicit directive in your response's text content:

**\textit{[STAGE\_SWITCH: perceive]}**

**If you need more retrieve tools, call them. ONLY output [STAGE\_SWITCH: perceive] when the retrieve goal (finding time segments) is met.**

\textbf{\textless RETRIEVE\_STAGE\_INSTRUCTION \textgreater}
\end{tcolorbox}
\caption{Instruction for Retrieve Phase.}
\label{prompt_retrieve}
\end{figure*}

\begin{figure*}[!t]
\begin{tcolorbox}
[
colframe=black!40!white,
colback=gray!10!white,
sharp corners=south,
title=\textbf{Instruction for Perceive Phase.}
]
\textbf{\textless PERCEPTION\_STAGE\_INSTRUCTION \textgreater}

**\textbf{Current Goal}:** Extract precise visual evidence from the video frames to confirm or deny the information gathered in the retrieve stage.

**CRITICAL RULE:** After locating a potential answer in the script/retrieve, you MUST use a perception tool to **CONFIRM** the visual evidence. This is the only stage where a final answer can be generated.

\vspace{1em}

**\textbf{Tool Usage Strategy (Forced Perception)}:**

\quad 1.  **Targeted Perception (Mandatory):** You MUST call at least one of the following tools: `object\_detect\_tool`, `boundary\_detect\_tool`, or `text\_extract\_tool` on the time ranges identified in the Retrieve stage to extract precise visual evidence.

\quad 2.  **Deep Dive (Last Resort):** **ONLY** if the information gathered from multiple calls to other perception tools is **consistently insufficient** to answer the question, you may call `frame\_analysis\_tool`. When using `frame\_analysis\_tool`, provide **a comprehensive list of all promising time segments** for the most detailed analysis.

\vspace{1em}

**\textbf{Stage Completion \& Switch}:**
Once you have located the relevant visual information necessary to proceed to inspect for output (Review Stage), you must include the following explicit directive in your response's text content:

**\textit{[STAGE\_SWITCH: review]}**

**If you need more perceive tools, call them. ONLY output [STAGE\_SWITCH: review] when the perceive goal is met.**

\textbf{\textless PERCEPTION\_STAGE\_INSTRUCTION \textgreater}
\end{tcolorbox}
\caption{Instruction for Perceive Phase.}
\label{prompt_perceive}
\end{figure*}

\begin{figure*}[!t]
\begin{tcolorbox}
[
colframe=black!40!white,
colback=gray!10!white,
sharp corners=south,
title=\textbf{Instruction for Review Phase.}
]

\textless \textbf{REVIEW\_STAGE\_INSTRUCTION} \textgreater

**\textbf{Current Goal}:** Review all textual and visual information with precise visual evidence to confirm if you can answer the question.

**\textbf{Incompletion \& Stage Switch}:**
If you need more visual information, return back to the Perception Stage, you must include the following explicit directive in your response's text content:
**\textit{[STAGE\_SWITCH: perceive]}**

\vspace{1em}

**\textbf{Task Completion Analysis \& Final Answer}:**

You must only provide a final answer if you have high confidence based on the gathered visual evidence.

**\textbf{Analysis Guidelines}:**

\quad - **Be conservative:** Only provide a final answer if the evidence is conclusive and visually confirmed.

**\textbf{Response Format}:**

\quad - If you can answer the question: Provide your final answer starting with `**Answer:**`

\quad - Multiple Choice Example: `**Answer:**A`

\quad - Time Localization Example: `**Answer:**[1.5s,12.5s]`.

\textless \textbf{REVIEW\_STAGE\_INSTRUCTION} \textgreater

\end{tcolorbox}
\caption{Instruction for Review Phase.}
\label{prompt_review}
\end{figure*}

% \section{Rationale}
% \label{sec:rationale}
% % 
% Having the supplementary compiled together with the main paper means that:
% % 
% \begin{itemize}
% \item The supplementary can back-reference sections of the main paper, for example, we can refer to \cref{sec:intro};
% \item The main paper can forward reference sub-sections within the supplementary explicitly (e.g. referring to a particular experiment); 
% \item When submitted to arXiv, the supplementary will already included at the end of the paper.
% \end{itemize}
% % 

% To split the supplementary pages from the main paper, you can use \href{https://support.apple.com/en-ca/guide/preview/prvw11793/mac#:~:text=Delete%20a%20page%20from%20a,or%20choose%20Edit%20%3E%20Delete).}{Preview (on macOS)}, \href{https://www.adobe.com/acrobat/how-to/delete-pages-from-pdf.html#:~:text=Choose%20%E2%80%9CTools%E2%80%9D%20%3E%20%E2%80%9COrganize,or%20pages%20from%20the%20file.}{Adobe Acrobat} (on all OSs), as well as \href{https://superuser.com/questions/517986/is-it-possible-to-delete-some-pages-of-a-pdf-document}{command line tools}.

\end{document}